\documentclass[letterpaper, 10 pt, conference]{ieeeconf}

\usepackage{amsfonts}
\usepackage{amsmath}
\usepackage{array}
\usepackage{bm}
\usepackage{booktabs}
\usepackage{caption}
\usepackage{eso-pic}
\usepackage{graphicx}
\usepackage{multirow}
\usepackage{subcaption}
\usepackage{tikz}
\usepackage[normalem]{ulem}
\usepackage{xcolor}
\usepackage{xspace}

\useunder{\uline}{\ul}{}

\makeatletter
\let\NAT@parse\undefined
\makeatother
\usepackage[colorlinks,urlcolor=blue,citecolor=red,bookmarks=false]{hyperref}

\IEEEoverridecommandlockouts                              

\newenvironment{stusubfig}[1]
{
	\begin{figure}[#1]
		\begin{center}
		}
		{
		\end{center}
	\end{figure}
}

\newenvironment{stusubfig*}[1]
{
	\begin{figure*}[#1]
		\begin{center}
		}
		{
		\end{center}
	\end{figure*}
}

\newcolumntype{H}{>{\setbox0=\hbox\bgroup}c<{\egroup}@{}}

\title{\LARGE \bf Sample, Crop, Track: Self-Supervised Mobile 3D Object Detection \\ for Urban Driving LiDAR}

\author{
\begin{tabular}{c@{\hskip 0.85cm}c@{\hskip 0.85cm}c@{\hskip 0.85cm}c}
Sangyun Shin & Stuart Golodetz & Madhu Vankadari & Kaichen Zhou \\
& Andrew Markham & Niki Trigoni
\end{tabular}%
\thanks{All authors are with the University of Oxford, UK.}
\thanks{E-mails: \texttt{\{firstname.lastname\}@cs.ox.ac.uk}}
\thanks{This work was supported by AWS via the Oxford-Singapore Human-}
\thanks{Machine Collaboration Programme.}
}

\begin{document}

\maketitle
\thispagestyle{empty}
\pagestyle{empty}

\begin{abstract}
\noindent Deep learning has led to great progress in the detection of mobile (i.e.\ movement-capable) objects in urban driving scenes in recent years. Supervised approaches typically require the annotation of large training sets; there has thus been great interest in leveraging weakly, semi- or self-supervised methods to avoid this, with much success. Whilst weakly and semi-supervised methods require some annotation, self-supervised methods have used cues such as motion to relieve the need for annotation altogether. However, a complete absence of annotation typically degrades their performance, and ambiguities that arise during motion grouping can inhibit their ability to find accurate object boundaries. In this paper, we propose a new self-supervised mobile object detection approach called SCT. This uses both motion cues and expected object sizes to improve detection performance, and predicts a dense grid of 3D oriented bounding boxes to improve object discovery. We significantly outperform the state-of-the-art self-supervised mobile object detection method TCR on the KITTI tracking benchmark, and achieve performance that is within 30\% of the fully supervised PV-RCNN++ method for IoUs $\le$ 0.5.
\end{abstract}

\section{Introduction}

Recent years have seen significant improvements in the accuracy and robustness of 3D object detectors, driven by the ever-improving capabilities of Deep Neural Networks (DNNs) and the commercial importance of tasks such as detecting vehicles, cyclists and pedestrians for autonomous driving. However, despite the success of DNNs, labelling the large datasets that fully supervised methods typically require has remained cumbersome and tedious.

Many recent semi-supervised \cite{wang20213dioumatch} and weakly-supervised \cite{zhu2019learning,choe2019attention,xue2019danet,choe2020evaluating,zhang2020inter,pan2021unveiling,gao2021ts} methods, which annotate only part of the dataset or use only high-level labels, have been proposed to reduce this burden. By contrast, self-supervised approaches are less common, owing to the difficulty of detecting objects without any annotation. One recent work that does use self-supervision is TCR \cite{harley2021track}, which groups nearby points with similar motions and applies sophisticated data augmentation to learn to detect mobile (i.e.\ movement-capable) objects in both 2D and 3D space. However, this approach risks generating inaccurate pseudo-ground truth boxes when e.g.\ two nearby objects move in the same way. Moreover, without a prior notion of the expected object sizes, noise in the estimated motions can lead points to be incorrectly included/excluded from an object, or clusters of background points with similar motions to be undesirably treated as objects of interest.

\begin{stusubfig}{!t}
	\begin{subfigure}{.49\linewidth}
		\centering
		\includegraphics[width=\linewidth]{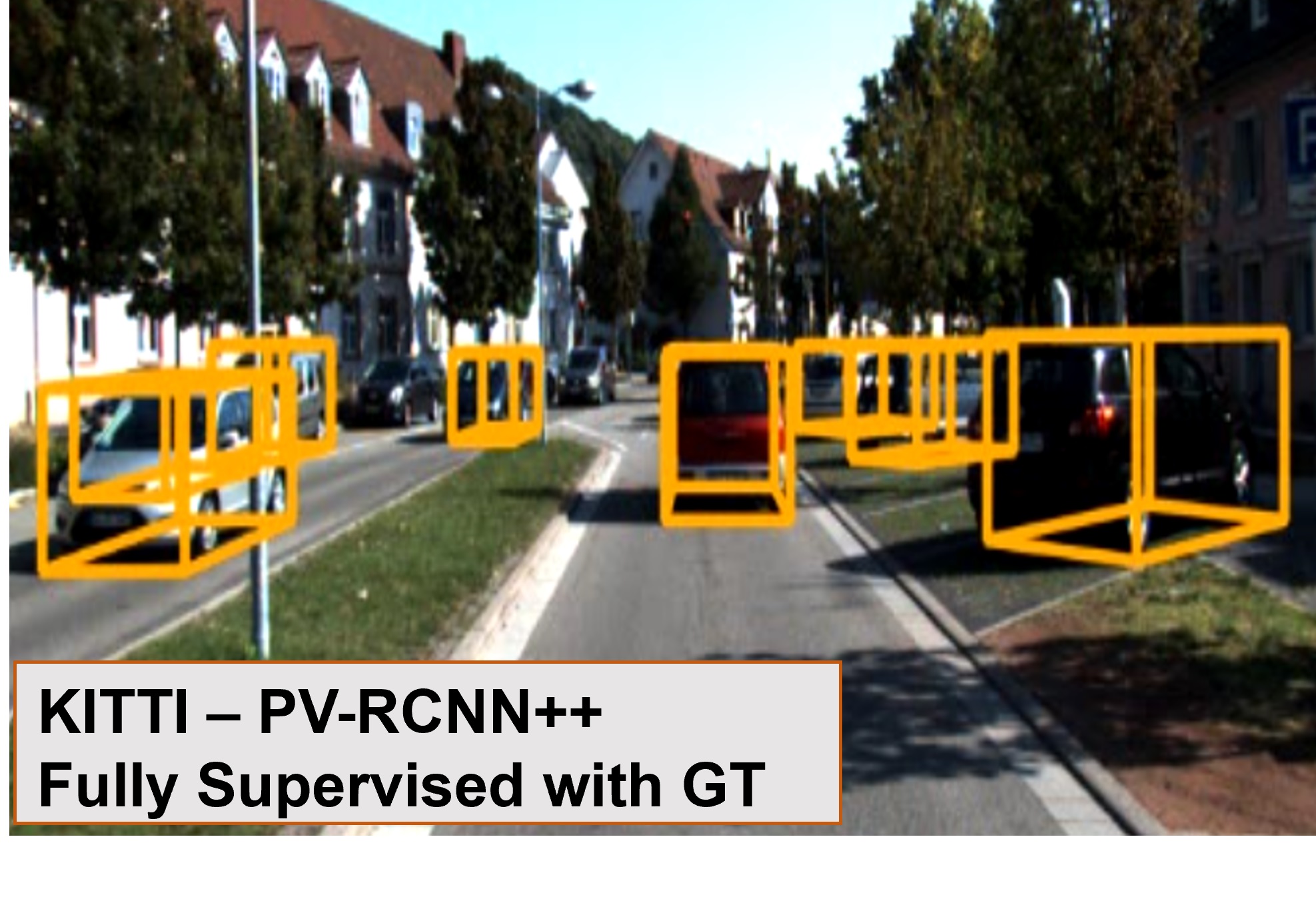}
	\end{subfigure}%
	\hfill%
	\begin{subfigure}{.49\linewidth}
		\centering
		\includegraphics[width=\linewidth]{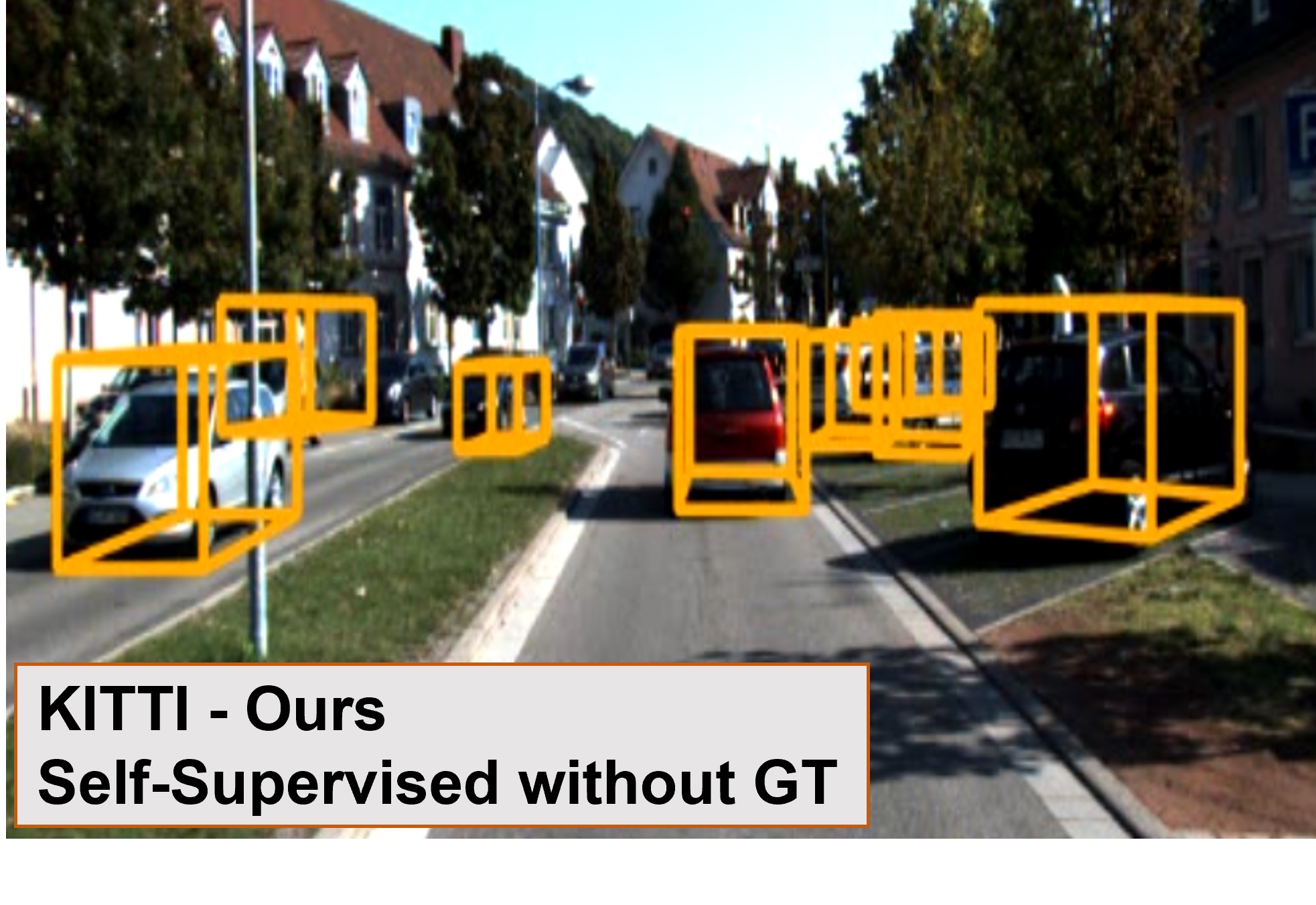}
	\end{subfigure}%
	\\[1mm]
	\begin{subfigure}{.49\linewidth}
		\centering
		\includegraphics[width=\linewidth]{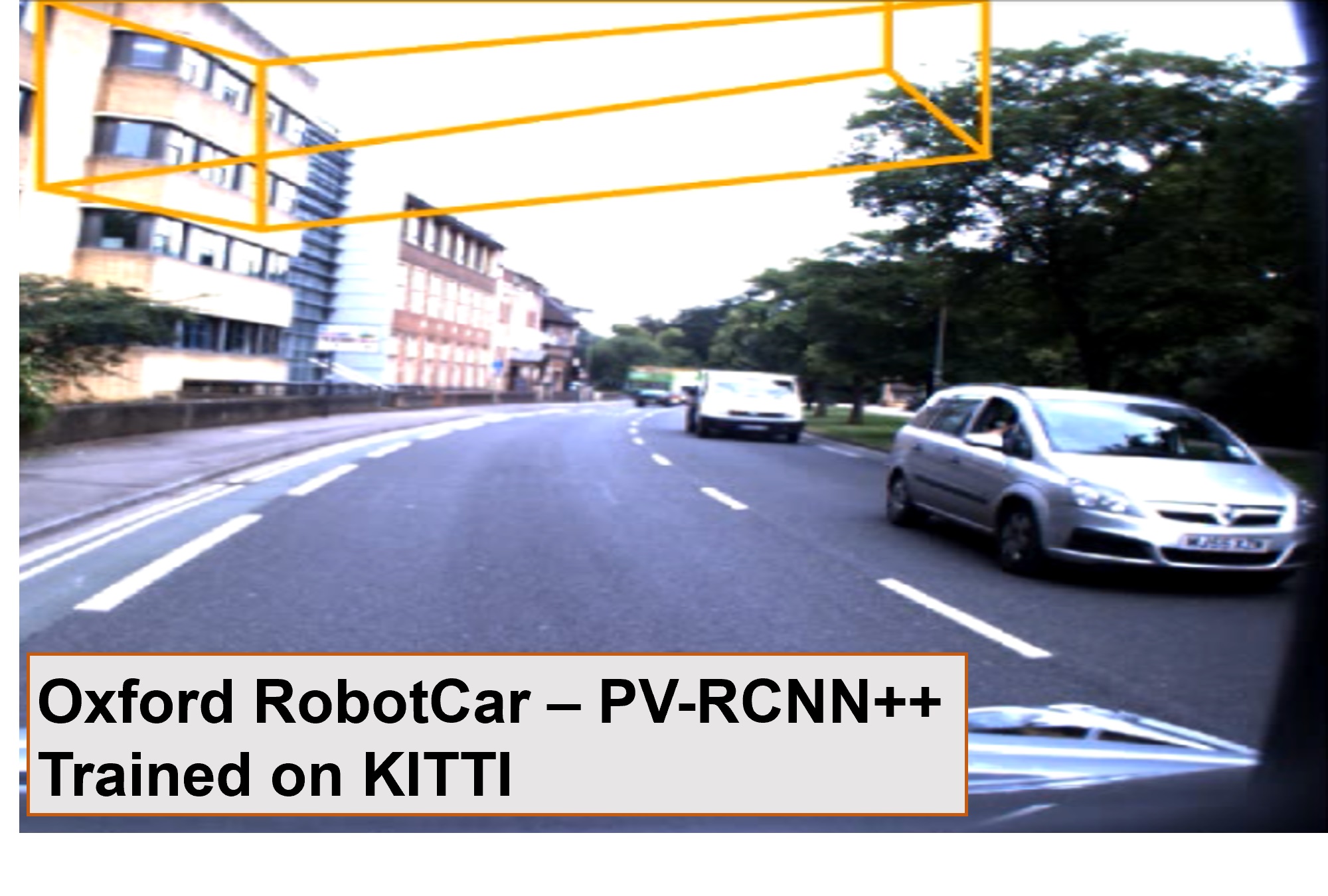}
	\end{subfigure}%
	\hfill%
	\begin{subfigure}{.49\linewidth}
		\centering
		\includegraphics[width=\linewidth]{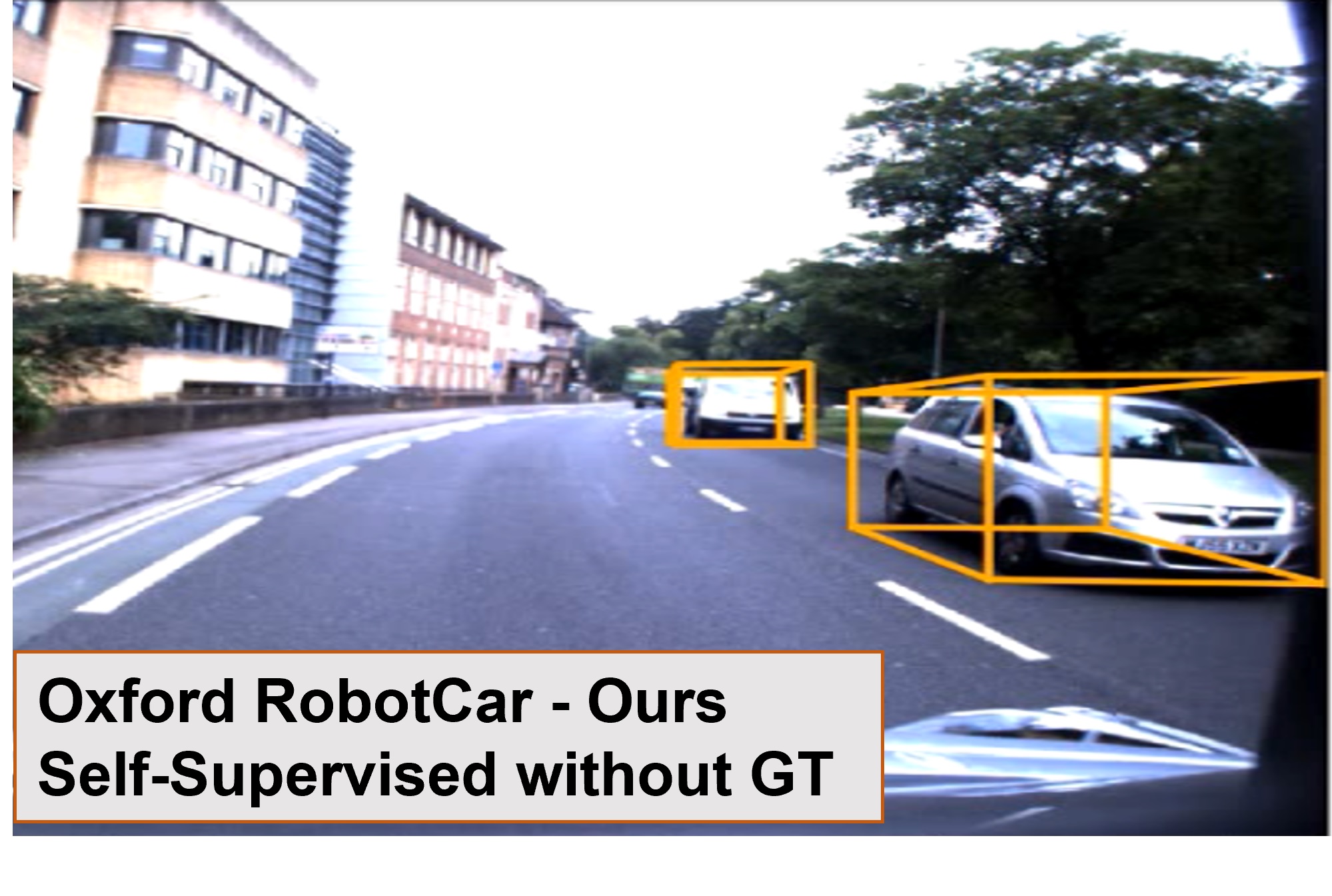}
	\end{subfigure}%
	\caption{Comparing our method with PV-RCNN++~\cite{shi2021pv} on the KITTI~\cite{geiger2012we} and Oxford RobotCar~\cite{RobotCarDatasetIJRR} datasets. For KITTI, where ground-truth annotations are available, the fully-supervised PV-RCNN++ outperforms our self-supervised approach. For RobotCar, no ground-truth annotations are available for training PV-RCNN++, and the KITTI-trained model fails to generalise to the different LiDAR setup used. By contrast, our self-supervised method does not use ground-truth annotations and can thus be trained on RobotCar, qualitatively outperforming PV-RCNN++ for this dataset.}
	\label{fig:teaser}
	\vspace{-1.5\baselineskip}
\end{stusubfig}

\begin{figure*}[!t]
	\centering
	\vspace{2mm}
	\includegraphics[width=.9\linewidth]{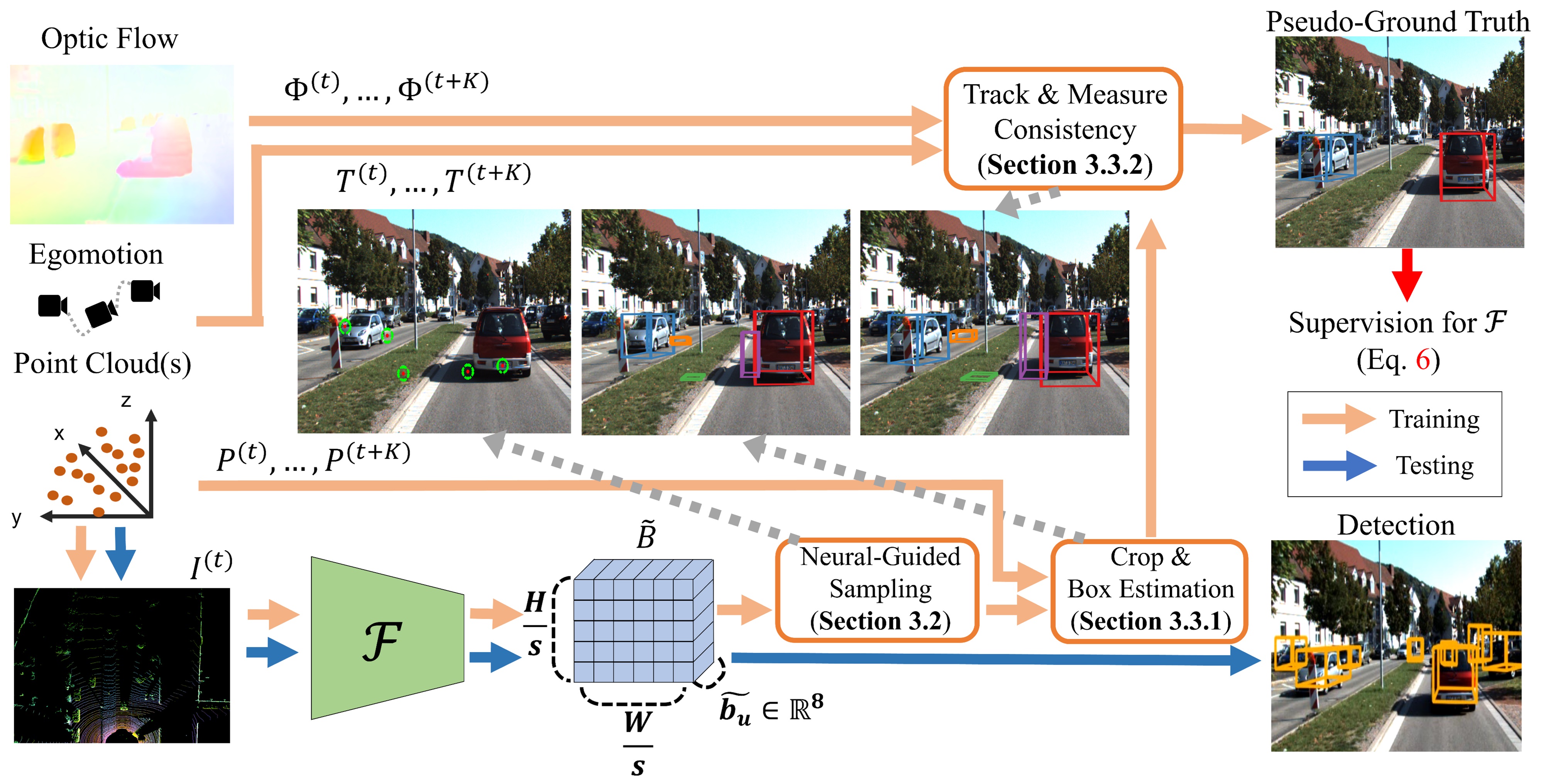}
	\caption{Our proposed approach (see \S\ref{Sec:method}). During training, we utilise cues from multiple successive frames. For inference, only a single birds-eye view (BEV) image is needed.}
	\label{fig:flow}
	\vspace{-\baselineskip}
\end{figure*}

In this paper, we address these issues by using the expected object sizes in the training set (which we require up-front) and temporal consistency in fitting boxes across multiple frames to filter out poor pseudo-ground truth boxes. By considering the expected object sizes when grouping points, we can avoid generating pseudo-ground truth boxes of inappropriate sizes, such as those that cover multiple nearby objects with similar motions. Considering the temporal consistency of fitted boxes across multiple frames helps to filter out further poor pseudo-ground truth boxes, such as those that are of an acceptable size, but overlap more than one object. Lastly, to generate accurate pseudo-ground truth boxes for closely located objects and objects that exhibit limited movement during training, we allow our model to predict a dense output image in which each pixel denotes a candidate box (this contrasts with TCR \cite{harley2021track}, which only considers a subset of the output pixels as candidates).

\textbf{Contributions:}
\textbf{1)} We use the expected sizes of objects in the training set and temporal consistency in fitting boxes across multiple frames to filter out poor pseudo-ground truth boxes.
\textbf{2)} We predict a dense output image in which each pixel denotes a candidate box, improving the accuracy of the pseudo-ground truth boxes our model can generate.
\textbf{3)} We propose a novel neural-guided sampling process to choose appropriate candidate pixels from the dense output image to both improve the efficiency of the training process, and guide the network towards dynamic objects during training.

\section{Related Work}

Many approaches have been proposed to tackle urban driving 3D object detection over the years. Our own method is entirely self-supervised, and so for brevity we focus only on the self-supervised literature in this paper. For recent reviews that survey the broader field, see \cite{arnold2019survey,wu2020deep}.

Early work in this area focused on grouping regions in an image to find if they were the same type as an object of interest~\cite{russell2006using,sivic2005discovering,weber2000towards}. The recent success in deep learning has led to many new works such as AIR~\cite{eslami2016attend}, MONet~\cite{burgess2019monet}, GENESIS~\cite{engelcke2019genesis}, IODINE~\cite{greff2019multi}, Slot Attention~\cite{locatello2020object} and SCALOR~\cite{Jiang*2020SCALOR:}. Most of these perform image reconstruction, in some cases using ``slots'' that are learned to represent an object or a part of an object using generative modelling. All these methods have proved to work well on simple datasets, but have struggled to handle the complexities of real-world data~\cite{harley2021track,bao2022discovering}. In very recent work, inspired by slot attention techniques, Bao et al.~\cite{bao2022discovering} used a weak motion segmentation algorithm to detect multiple mobile objects in synthesised photorealistic 2D images; however, the KITTI training code and models for this approach have not been publicly released.

Another line of research uses motion cues to detect objects, assuming that the regions corresponding to each object should share the same motion. The early works in this direction combined the affinity matrix from point trajectories based on 2D motions in a video and standard clustering techniques to detect objects~\cite{Bro10c,kiadaki2012V}. These works assumed the camera to be static. In a more recent work, Chen et al.~\cite{chen2021moving} used range images from a moving camera and processed them through CNNs to detect moving regions by labelling range data. In another recent work, \emph{Track, Check, Repeat}~\cite{harley2021track} (TCR) proposed an Expectation Maximisation (EM) approach for mobile object detection in the context of autonomous driving by using other information such as ego-motion, 2D optic flow and 3D LiDAR data.

\section{Method}
\label{Sec:method}

\subsection{Overview}
\label{Sec:overview}

\begin{figure*}[!t]
	\centering
	\vspace{2mm}
	\includegraphics[width=.7\linewidth]{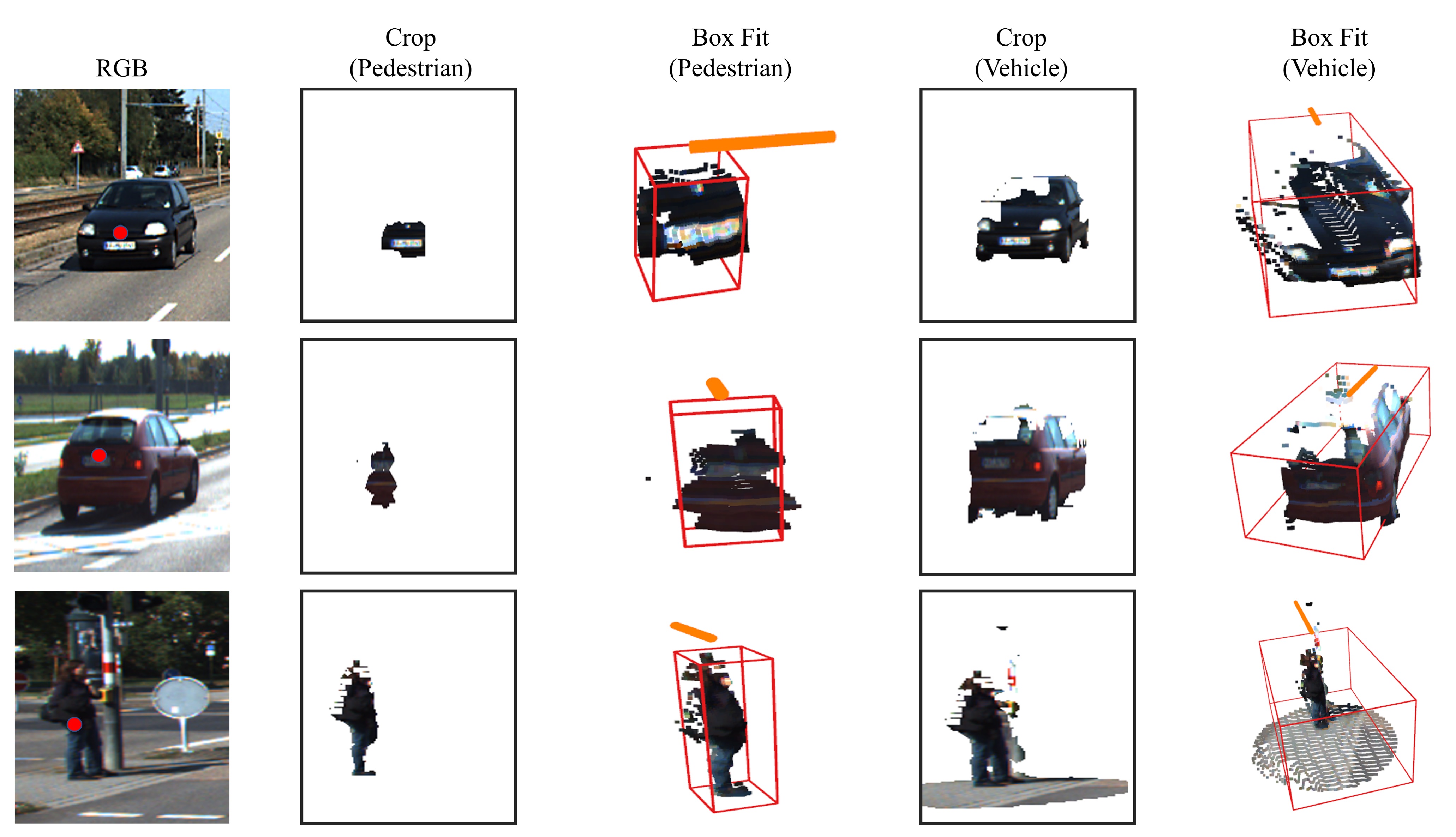}
	\caption{Cropping and box fitting examples. The red dot on each RGB image indicates a chosen pixel $\mathbf{u}$ in $\tilde{B}^{(t)}$. As per \S\ref{subsec:cropping}, we crop a set of points from $P^{(t)}$ around the centre $\tilde{\mathbf{c}}_\mathbf{u}^{(t)}$ of the bounding box denoted by $\mathbf{u}$, for each anchor, and then fit a 3D oriented bounding box to the cropped points. The pedestrian anchor is too small to capture the entirety of the cars, whereas the vehicle anchor performs far better; the converse is true for the pedestrian example, where the vehicle anchor captures too much background. Orange lines show the orientations of the fitted boxes.}
	\label{fig:scan}
	\vspace{-\baselineskip}
\end{figure*}

Our method uses self-supervision to train a model $\mathcal{F}$\footnote{Our $\mathcal{F}$ uses an encoder-decoder architecture, in which the encoder is based on Feature Pyramid Network (FPN)~\cite{lin2017feature} and ResNet-18~\cite{he2016deep}, and the decoder is based on CenterNet~\cite{duan2019centernet} (without any pre-trained weights).} that can predict oriented bounding boxes around mobile objects in a 3D urban driving scene (see Figure~\ref{fig:flow}). More specifically, we discretise a 3D volume in front of the ego-vehicle using an $H \times W$ grid of vertical pillars, and aim to detect objects that fall within this volume. The input to $\mathcal{F}$ is a birds-eye view (BEV) representation $I \in \mathbb{R}^{H \times W \times 3}$ of a 3D LiDAR point cloud $P$. This is formed in the standard KITTI way \cite{geiger2012we} by projecting $P$ down into the plane, such that each pixel in $I$ then corresponds to a single pillar in the grid.\footnote{In practice, $H = W = 608$. The grid is axis-aligned, and corresponds to a 3D volume in LiDAR space covering points $(x,y,z) \in P$ that satisfy $\mbox{2.5m} \le x \le \mbox{40m}$, $\mbox{-18m} \le y \le \mbox{18m}$ and $\mbox{-2.73m} \le z \le \mbox{1.27m}$. The three values for each \{pixel/grid cell\} are based on the heights, intensities, and density of the points from $P$ in the corresponding grid cell.} The output of $\mathcal{F}$ is a (downsampled) image $\tilde{B} \in \mathbb{R}^{\frac{H}{s} \times \frac{W}{s} \times 8}$, in which $s = 4$ is the downsampling factor. Each pixel $\mathbf{u}$ in $\tilde{B}$ contains an $8$-tuple $\tilde{\mathbf{b}}_\mathbf{u} = (\tilde{\bm{\delta}}_\mathbf{u}, \tilde{\mathbf{d}}_\mathbf{u}, \tilde{r}_\mathbf{u}, \tilde{\kappa}_\mathbf{u})$ denoting a predicted 3D bounding box in KITTI~\cite{geiger2012we} format. In this, $\tilde{\mathbf{d}}_\mathbf{u} \in \mathbb{R}^3$ denotes the (axis-aligned) dimensions of the box, $\tilde{r}_\mathbf{u} \in [-\pi,\pi]$ its rotation angle (yaw), and $\tilde{\kappa}_\mathbf{u} \in [0,1]$ its confidence score. The centre $\tilde{\mathbf{c}}_\mathbf{u}$ of the bounding box is specified indirectly as an offset $\tilde{\bm{\delta}}_\mathbf{u} \in \mathbb{R}^3$ from the centre of the vertical pillar in the grid that corresponds to $\mathbf{u}$.\footnote{This is important, as it implicitly constrains the centres (indirectly) predicted by the network at the start of training to be within a reasonable distance of the relevant pillars.}

Note that at training time, our method also requires access to sparse depth images $\{D^{(t)} \in \mathbb{R}^{H' \times W'}\}$, optic flow images $\{\Phi^{(t)} \in \mathbb{R}^{H' \times W' \times 2}\}$ and 6DoF poses $\{T_t^w \in \mathbb{SE}(3)\}$, expressed as transformations from camera space at frame $t$ to world space $w$. These are typically available (or easily obtained) in the urban driving setting on which we focus.\footnote{We constructed each $D^{(t)}$ by projecting the points in the corresponding point cloud $P^{(t)}$ down into the image plane (using first the fixed, pre-calibrated transformation $S_L^C$ from LiDAR to camera space, and then perspective projection based on the known camera intrinsics). To obtain the optic flow images, we followed TCR \cite{harley2021track} in training a RAFT \cite{teed2020raft} model on MPI Sintel Flow \cite{butler2012sintel}. For our experiments on the KITTI tracking benchmark \cite{geiger2012we}, $H'$ was $375$ and $W'$ was $1242$. We computed the 6DoF poses from the OxTS GPS/IMU data provided by the benchmark.}

We draw inspiration from a recent study~\cite{harley2021track} that showcased the possibility of training a model using only the labels of actively moving objects and appropriate data augmentation to detect mobile objects with a similar shape, regardless of whether or not they are moving at the time. In our case, the set of shapes we expect objects to have must be specified in advance as a set of anchors $A$. Each anchor $\mathbf{a} = (a_x, a_y, a_z) \in A$ denotes an axis-aligned bounding box (AABB) of size $a_x \times a_y \times a_z$.\footnote{We used three anchors: pedestrian, cyclist and vehicle. Our pedestrian anchor $\mathbf{a}_p = (0.45\mbox{m}, 1.70\mbox{m}, 0.27\mbox{m})$ was obtained from \cite{iso_2015_7250_3}. Our cyclist anchor $\mathbf{a}_c = (0.54\mbox{m}, 1.90\mbox{m}, 1.75\mbox{m})$ was derived by adding half the height of the pedestrian anchor to the average dimensions from \cite{bestbikelock}. Our vehicle anchor $\mathbf{a}_v = (1.88\mbox{m}, 1.63\mbox{m}, 4.58\mbox{m})$ was obtained from \cite{nimblefins}.} To train $\mathcal{F}$ to detect mobile objects without manual annotation, we propose a multi-step process that is run for each frame $(P^{(t)}, I^{(t)})$ in the training set to generate pseudo-ground truth 3D bounding boxes that can be used to train our model. First, we compute $\tilde{B}^{(t)} = \mathcal{F}(I^{(t)})$. Next, we use a neural-guided sampling process (see \S\ref{subsec:sampling}) to choose a subset of the pixels from $\tilde{B}^{(t)}$ to use for self-supervision. Each chosen pixel $\mathbf{u}$ denotes a predicted bounding box $\tilde{\mathbf{b}}_\mathbf{u}^{(t)}$. We then aim to generate (see \S\ref{subsec:pseudogt}) a pseudo-ground truth box $\mathbf{b}_\mathbf{u}^{(t)}$ corresponding to each $\tilde{\mathbf{b}}_\mathbf{u}^{(t)}$ that can be compared to it as part of the loss (see \S\ref{subsec:loss}). We now describe this entire process in more detail.

\subsection{Sampling}
\label{subsec:sampling}

As mentioned in \S\ref{Sec:overview}, the overall goal of our self-supervised approach is to generate pseudo-ground truth boxes for a subset of the pixels in $\tilde{B}^{(t)}$, which can then be used to supervise the training of our model. Empirically, we found generating a pseudo-ground truth box for every pixel in $\tilde{B}^{(t)}$ to be prohibitively costly ($\approx{}1$ hour/frame); we thus propose a neural-guided approach to choose a subset of the pixels in $\tilde{B}^{(t)}$ to use for self-supervision.
Our approach aims to select a mixture of some pixels whose boxes have high confidence scores predicted by the model, and some whose boxes have lower confidence scores.\footnotemark{} To achieve this, we first use a threshold $\tau = 0.08$ to divide the pixels into two sets -- those whose confidence $\kappa_\mathbf{u}^{(t)} > \tau$, and those whose confidence $\kappa_\mathbf{u}^{(t)} \le \tau$ -- and then uniformly sample $N/2 = 30$ pixels from each set. Rather than directly using the confidences predicted by the network for these boxes, we then compute a more robust confidence for each box by averaging the confidences of the box itself and the $8$ boxes from $\tilde{B}^{(t)}$ whose centres are the nearest neighbours of the box's centre in 3D. This makes the confidence values for the boxes more robust to local noise.

\footnotetext{Sampling high-confidence boxes allows us to continually learn how to predict boxes for the same objects over time; conversely, sampling low-confidence boxes allows us to discover new objects in the scene.}

\subsection{Pseudo-Ground Truth Box Generation}
\label{subsec:pseudogt}

Each pixel $\mathbf{u}$ from $\tilde{B}^{(t)}$ chosen as described in \S\ref{subsec:sampling} denotes a predicted bounding box $\tilde{\mathbf{b}}_\mathbf{u}^{(t)}$ centred on the 3D point $\tilde{\mathbf{c}}_\mathbf{u}^{(t)}$. For each such $\mathbf{u}$, we now seek to generate a corresponding pseudo-ground truth box $\mathbf{b}_\mathbf{u}^{(t)}$ that can be compared to $\tilde{\mathbf{b}}_\mathbf{u}^{(t)}$ as part of the loss (see \S\ref{subsec:loss}). The sizes of the pseudo-ground truth boxes we generate will be guided by the initial set of anchors (expected object sizes) provided by the user (see \S\ref{subsec:cropping}).

To generate the boxes, we proceed as follows. For each $\mathbf{u}$ and each anchor $\mathbf{a}$, we crop a set of points $Q_{\mathbf{u},\mathbf{a}}^{(t)}$ from $P^{(t)}$ around $\tilde{\mathbf{c}}_\mathbf{u}^{(t)}$ (see \S\ref{subsec:cropping}). We then track the points in $Q_{\mathbf{u},\mathbf{a}}^{(t)}$ forwards for $K$ frames, resulting in $K+1$ sets of points $\{Q_{\mathbf{u},\mathbf{a}}^{(t\rightsquigarrow{}k)} : 0 \le k \le K\}$ for each $\mathbf{u}$ and $\mathbf{a}$ (see \S\ref{subsec:tracking}). Next, we fit a 3D oriented bounding box (OBB) to each $Q_{\mathbf{u},\mathbf{a}}^{(t\rightsquigarrow{}k)}$, and then use these boxes to choose the best anchor $\mathbf{a}^\star$ (if any) for each $\mathbf{u}$. For any pixel $\mathbf{u}$ with a valid best anchor $\mathbf{a}^\star$, we treat the box fitted to $Q_{\mathbf{u},\mathbf{a}^\star}^{(t\rightsquigarrow{}0)}$ as the pseudo-ground truth for $\mathbf{u}$, convert it into the correct space and denote it as $\mathbf{b}_\mathbf{u}^{(t)}$ (see \S\ref{subsec:boxselection}).

\subsubsection{Cropping}
\label{subsec:cropping}

For a chosen pixel $\mathbf{u}$ in $\tilde{B}^{(t)}$, denoting a predicted box with centre $\tilde{\mathbf{c}}_\mathbf{u}^{(t)} = (c_x, c_y, c_z)$, we first crop a set of points $Q_{\mathbf{u},\mathbf{a}}^{(t)}$ from $P^{(t)}$ in a vertical cylinder\footnote{Thus obviating the need to know the yaw of the object \emph{a priori}.} around $\tilde{\mathbf{c}}_\mathbf{u}^{(t)}$ for each user-provided anchor $\mathbf{a} = (a_x, a_y, a_z) \in A$. The points in $P^{(t)}$ are in the space defined by the LiDAR at frame $t$; for convenience, we convert them into camera space at frame $t$ by applying the (fixed, pre-calibrated) transformation $S_L^C \in \mathbb{SE}(3)$ from LiDAR to camera space. Formally, $Q_{\mathbf{u},\mathbf{a}}^{(t)}$ can then be defined as
\begin{equation}
\footnotesize
\begin{aligned}
Q_{\mathbf{u},\mathbf{a}}^{(t)} = & \left\{ S_L^C \mathbf{p} : \mathbf{p} = (p_x,p_y,p_z) \in P^{(t)} \right. \\
& \; \left. \mbox{ and } |p_y - c_y| < \frac{a_y}{2} \right. \\
& \; \left. \mbox{ and } \left\| (p_x - c_x, p_z - c_z) \right\|_2 < \frac{\left\| \left(a_x, a_z\right) \right\|_2}{2} \right\}.
\end{aligned}
\end{equation}

\subsubsection{Tracking}
\label{subsec:tracking}

We next track the points in $Q_{\mathbf{u},\mathbf{a}}^{(t)}$ forwards for $K$ frames to help us to determine the best anchor (if any) for each $\mathbf{u}$ (see \S\ref{subsec:boxselection}). We denote the point track for an individual point $\mathbf{q} \in Q_{\mathbf{u},\mathbf{a}}^{(t)}$ as $\{\chi_0(\mathbf{q}), \ldots, \chi_K(\mathbf{q})\}$. To calculate it, we first set $\chi_0(\mathbf{q})$ to $\mathbf{q}$, and then iteratively compute each $\chi_{k+1}(\mathbf{q})$ from the preceding $\chi_k(\mathbf{q})$ as follows. First we project $\chi_k(\mathbf{q})$, which is in camera space, down into the image plane via $\mathbf{u}_k = \pi(\chi_k(\mathbf{q}))$, in which $\pi$ denotes perspective projection. Then we compute the corresponding pixel in frame $t+k+1$, namely $\mathbf{u}'_{k+1} = \mathbf{u}_k + \Phi^{(t+k)}$, find the nearest pixel $\mathbf{u}''_{k+1}$ to it that has a valid depth in $D^{(t+k+1)}$, and back-project this into camera space to obtain $\chi_{k+1}(\mathbf{q}) = \pi^{-1}\left(\mathbf{u}''_{k+1}, D^{(t+k+1)}\right)$.\footnote{This approach can in some cases yield a $\chi_{k+1}(\mathbf{q})$ that is not a good match for $\chi_k(\mathbf{q})$, e.g.\ if $\mathbf{u}''_{k+1}$ is far from $\mathbf{u}'_{k+1}$. However, since only sets of boxes based on correct point tracks will survive the consistency scoring in \S\ref{subsec:boxselection}, we did not observe this to cause a problem in practice.} We then use the point tracks for all the points in $Q_{\mathbf{u},\mathbf{a}}^{(t)}$ to project $Q_{\mathbf{u},\mathbf{a}}^{(t)}$ forwards into the $K$ subsequent frames, via
\begin{equation}
\footnotesize
Q_{\mathbf{u},\mathbf{a}}^{(t\rightsquigarrow{}k)} =
\begin{cases}
    Q_{\mathbf{u},\mathbf{a}}^{(t)} & \mbox{if } k = 0, \\
    \{T_{t+k}^t \chi_k(\mathbf{q}) : \mathbf{q} \in Q_{\mathbf{u},\mathbf{a}}^{(t)}\} & \mbox{if } 0 < k \le K.
\end{cases}
\end{equation}
Note that we transform all of the points into frame $t$ here using $T_{t+k}^t = (T_t^w)^{-1} T_{t+k}^w$, since we want all of the $Q_{\mathbf{u},\mathbf{a}}^{(t\rightsquigarrow{}k)}$ point sets to be in the same frame of reference.

\begin{stusubfig}{!t}
    \vspace{2mm}
	\begin{subfigure}{.49\linewidth}
		\centering
		\includegraphics[width=\linewidth]{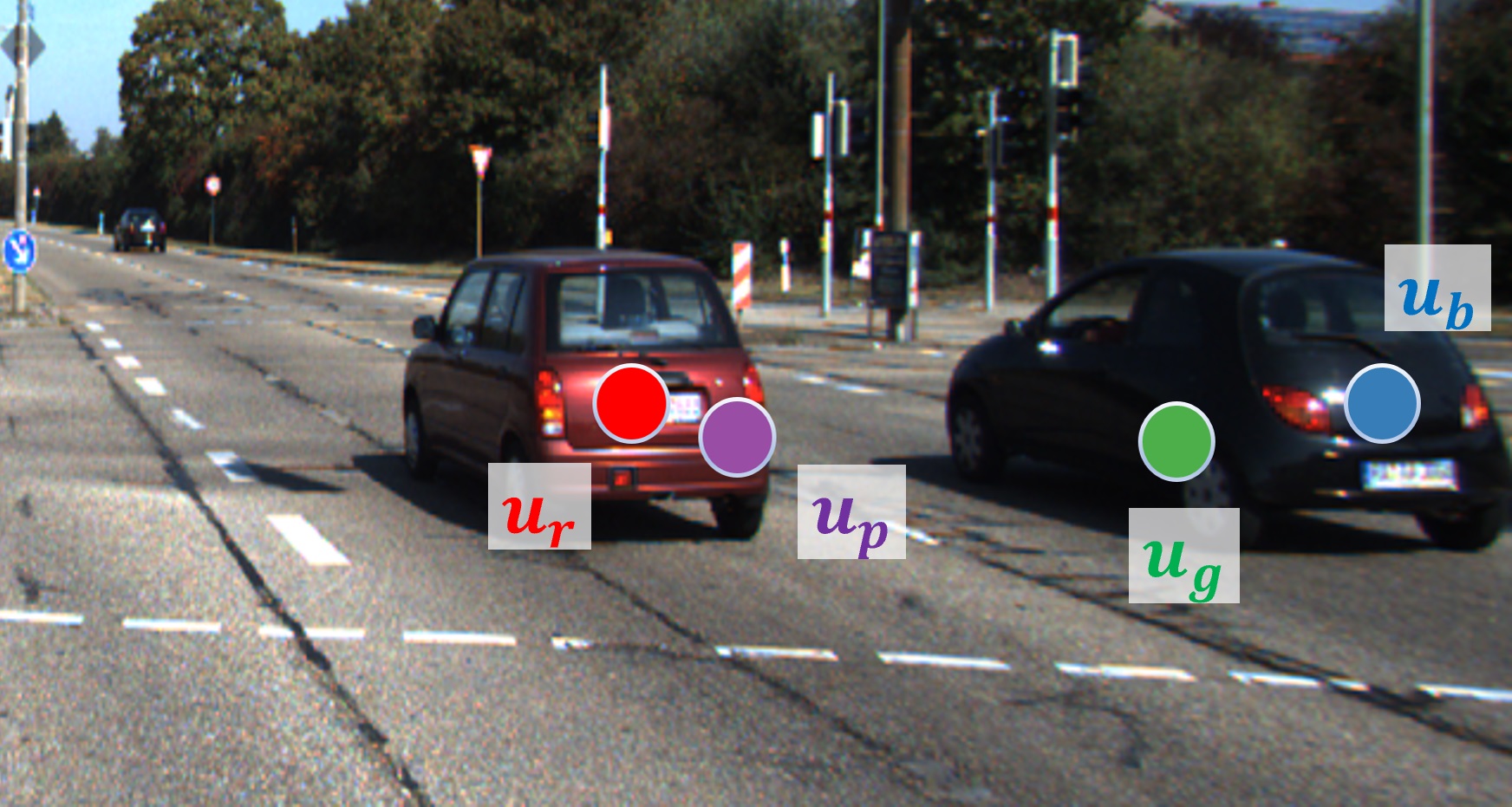}
	\end{subfigure}%
	\hfill%
	\begin{subfigure}{.49\linewidth}
		\centering
		\includegraphics[width=\linewidth]{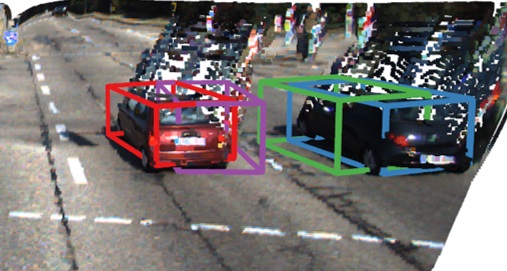}
	\end{subfigure}%
	\\[1mm]
	\begin{subfigure}{.49\linewidth}
		\centering
		\includegraphics[width=\linewidth]{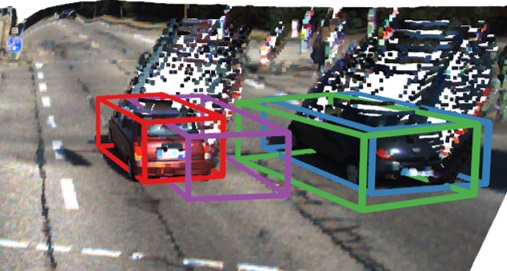}
	\end{subfigure}%
	\hfill%
	\begin{subfigure}{.49\linewidth}
		\centering
		\includegraphics[width=\linewidth]{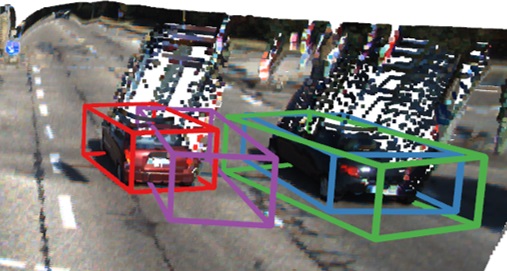}
	\end{subfigure}%
	\vspace{3mm}
	\caption{The use of temporal consistency to decide between pseudo-ground truth box candidates for objects. Left-to-right, top-to-bottom: choosing candidate pixels $U = \{\mathbf{u}_r,\mathbf{u}_p,\mathbf{u}_g,\mathbf{u}_b\}$; the fitted boxes $\mathbf{b}_{\mathbf{u},\mathbf{a}_v}^{(t\rightsquigarrow{}k)}$ for $\mathbf{u} \in U$, the vehicle anchor $\mathbf{a}_v$ and $0 < k \le 3$. The shapes of the red and blue boxes are more consistent over time than those of the purple and green ones, and so $\mathbf{b}_{\mathbf{u}_r}^{(t)}$ and $\mathbf{b}_{\mathbf{u}_b}^{(t)}$ will be used as the pseudo-ground truths for the cars.}
	\label{fig:tracking}
	\vspace{-\baselineskip}
\end{stusubfig}

\subsubsection{Box Fitting and Selection}
\label{subsec:boxselection}

Having constructed the point sets as described in \S\ref{subsec:tracking}, we next fit to each point set $Q_{\mathbf{u},\mathbf{a}}^{(t\rightsquigarrow{}k)}$ a 3D oriented bounding box $\mathbf{b}_{\mathbf{u},\mathbf{a}}^{(t\rightsquigarrow{}k)}$ that has centre $\mathbf{c} \in \mathbb{R}^3$, dimensions $\mathbf{d} \in \mathbb{R}^3$ and rotation angle (yaw) $r \in [-\pi,\pi]$. We compute $\mathbf{c}$ as the mean of the points in $Q_{\mathbf{u},\mathbf{a}}^{(t\rightsquigarrow{}k)}$. To find $r$, we first apply Principal Component Analysis (PCA) to $Q_{\mathbf{u},\mathbf{a}}^{(t\rightsquigarrow{}k)}$, and let $\mathbf{e} = (e_x,e_y,e_z) \in \mathbb{R}^3$ denote its first eigenvector. We then compute $r = \arctan(e_z/e_x)$, as our camera coordinate system has $x$ right, $y$ down and $z$ forward. We approximate the desired dimensions $\mathbf{d}$ as those of the smallest axis-aligned bounding box (AABB) that encloses the set of points $S = \{\mathcal{T}\mathbf{q} : \mathbf{q} \in Q_{\mathbf{u},\mathbf{a}}^{(t\rightsquigarrow{}k)}\}$, where $\mathcal{T} \in \mathbb{SE}(3)$ denotes a rigid transformation that translates $\mathbf{c}$ to the origin and rotates by $-r$ around the $y$ axis. Then $\mathbf{d}$ can be trivially calculated by first defining $\mathit{min}_i = \min_{\mathbf{s} \in S} s_i$ and $\mathit{max}_i = \max_{\mathbf{s} \in S} s_i$ for all $i \in \{x,y,z\}$, where $s_i$ denotes the $i^\mathit{th}$ component of $\mathbf{s}$, and then computing $\mathbf{d} = (\mathit{max}_x-\mathit{min}_x,\mathit{max}_y-\mathit{min}_y,\mathit{max}_z-\mathit{min}_z)$.

For each chosen pixel $\mathbf{u} \in \tilde{B}^{(t)}$, we will then have constructed a set of 3D oriented bounding boxes $\{\mathbf{b}_{\mathbf{u},\mathbf{a}}^{(t\rightsquigarrow{}k)} : 0 \le k \le K\}$ for each anchor $\mathbf{a} \in A$. We now seek to choose the best anchor $\mathbf{a}^\star$ (if any) to associate with each $\mathbf{u}$, which will in turn allow us to define a pseudo-ground truth bounding box $\mathbf{b}_\mathbf{u}^{(t)}$ to associate with $\mathbf{u}$ for training purposes. We will denote the set of chosen pixels for which best anchors can be found as $U_+^{(t)}$, and the remaining chosen pixels as $U_-^{(t)}$. To determine $\mathbf{b}_\mathbf{u}^{(t)}$, we initially score the set of boxes associated with each $\mathbf{u}$ and $\mathbf{a}$ using two criteria: (i) the extent to which the set of boxes indicates a moving object, and (ii) the extent to which the dimensions of the boxes in the set remain consistent over time. (Note that we want to generate pseudo-ground truth boxes that correspond to moving objects and are temporally consistent.) The first criterion can be specified as
\begin{equation}
\footnotesize
\mathit{moving}_{\mathbf{u},\mathbf{a}}^{(t)} = \sum_{k=1}^K \left\| \mathbf{c}_{\mathbf{u},\mathbf{a}}^{(t\rightsquigarrow{}k)} - \mathbf{c}_{\mathbf{u},\mathbf{a}}^{(t\rightsquigarrow{}(k-1))} \right\|_2,
\end{equation}
in which $\mathbf{c}_{\mathbf{u},\mathbf{a}}^{(t\rightsquigarrow{}k)}$ denotes the centre of $\mathbf{b}_{\mathbf{u},\mathbf{a}}^{(t\rightsquigarrow{}k)}$. This computes the distance moved by the object over the $K+1$ frames considered. The second criterion can be specified as
\begin{equation}
\footnotesize
\mathit{inconsistency}_{\mathbf{u},\mathbf{a}}^{(t)} = \sum_{k=1}^K \left\| \mathbf{d}_{\mathbf{u},\mathbf{a}}^{(t\rightsquigarrow{}k)} - \mathbf{d}_{\mathbf{u},\mathbf{a}}^{(t\rightsquigarrow{}0)} \right\|_2,
\end{equation}
in which $\mathbf{d}_{\mathbf{u},\mathbf{a}}^{(t\rightsquigarrow{}k)}$ denotes the dimensions of $\mathbf{b}_{\mathbf{u},\mathbf{a}}^{(t\rightsquigarrow{}k)}$. This sums variations from the initial box in the set. We combine both criteria into a single confidence score for each set of boxes via
\begin{equation}
\footnotesize
\kappa_{\mathbf{u},\mathbf{a}}^{(t)} = \lambda_1 \times \mathit{moving}_{\mathbf{u},\mathbf{a}}^{(t)} - \lambda_2 \times \mathit{inconsistency}_{\mathbf{u},\mathbf{a}}^{(t)},
\end{equation}
empirically setting $\lambda_1 = 0.4$ and $\lambda_2 = 0.15$.

\begin{table}[!t]
	\centering
	\vspace{2mm}
	\scriptsize
	\resizebox{\linewidth}{!}{%
		\begin{tabular}{lcccccccc}
			& \textbf{Eval.} & \multicolumn{7}{c}{\textbf{mAP@IoU}} \\
			&& \emph{0.1} & \emph{0.2} & \emph{0.3} & \emph{0.4} & \emph{0.5} & \emph{0.6} & \emph{0.7} \\
			\midrule
			\textbf{Self-Supervised} &&&&&&&& \\
			\midrule
			Slot Attention~\cite{locatello2020object} & 2D & 0.07 & 0.03 & 0.01 & 0.00 & 0.00 & 0.00 & 0.00 \\
			SCALOR~\cite{Jiang*2020SCALOR:} & 2D & 0.11 & 0.07 & 0.02 & 0.00 & 0.00 & 0.00 & 0.00 \\
			TCR~\cite{harley2021track} & 2D & 0.43 & 0.40 & 0.37 & 0.35 & 0.33 & 0.30 & 0.21 \\
			TCR$^\star$~\cite{harley2021track} & 2D & 0.60 & 0.59 & 0.59 & 0.56 & 0.49 & 0.40 & 0.23 \\
			Ours & 2D & \textbf{0.79} & \textbf{0.77} & \textbf{0.74} & \textbf{0.72} & \textbf{0.66} & \textbf{0.59} & \textbf{0.41} \\
			\midrule
			TCR~\cite{harley2021track} & BEV & 0.40 & 0.38 & 0.35 & 0.33 & 0.31 & 0.23 & 0.06 \\
			TCR$^\star$~\cite{harley2021track} & BEV & 0.58 & 0.58 & 0.54 & 0.48 & 0.42 & 0.29 & 0.10 \\
			Ours & BEV & \textbf{0.79} & \textbf{0.79} & \textbf{0.76} & \textbf{0.74} & \textbf{0.69} & \textbf{0.62} & \textbf{0.33} \\
			\midrule
			\textbf{Semi-Supervised} &&&&&&&& \\
			\midrule
			\emph{3DIoUMatch~\cite{wang20213dioumatch}} & \emph{BEV} & \emph{0.84} & \emph{0.84} & \emph{0.84} & \emph{0.84} & \emph{0.84} & \emph{0.84} & \emph{0.83} \\
			\midrule
			\textbf{Fully Supervised} &&&&&&&& \\
			\midrule
			\emph{PV-RCNN++~\cite{shi2021pv}} & \emph{BEV} & \emph{0.98} & \emph{0.98} & \emph{0.98} & \emph{0.97} & \emph{0.97} & \emph{0.95} & \emph{0.95} \\
			\bottomrule
		\end{tabular}
	}
	\vspace{\baselineskip}
	\caption{Comparing our method's ability to discover mobile objects to that of other methods on the KITTI tracking benchmark \cite{geiger2012we}. We compare to a number of recent methods, of which three are self-supervised, one is semi-supervised (with annotations for 10\% of the dataset) and one is fully supervised. TCR denotes the original version described in \cite{harley2021track}; TCR$^\star$ denotes the improved version published on GitHub. (We italicise the semi-supervised and fully supervised methods, which are not directly comparable as their access to annotations gives them a significant advantage.)}
	\label{tab:obj_discovery}
	\vspace{-\baselineskip}
\end{table}

To try to choose a best anchor $\mathbf{a}^\star$ for $\mathbf{u}$, we now first disregard any anchor $\mathbf{a}$ such that $\kappa_{\mathbf{u},\mathbf{a}}^{(t)} < \eta$, where $\eta$ is a threshold empirically set to $0.08$. We then examine the number of surviving candidate anchors for $\mathbf{u}$: if no anchors have survived, we add $\mathbf{u}$ to $U_-^{(t)}$; otherwise, if at least one anchor has survived, we add $\mathbf{u}$ to $U_+^{(t)}$. If exactly one anchor has survived, this is chosen as the best anchor; if multiple anchors have survived, we select the one with the largest 3D volume, on the basis that whilst smaller anchors may yield slightly higher confidences, owing to the lesser amount of background they contain, the largest anchor has met the threshold and is the one most likely to be correct in practice.

We can now define the pseudo-ground truth bounding box $\mathbf{b}_\mathbf{u}^{(t)}$ for each pixel $\mathbf{u} \in U_+^{(t)}$ with best anchor $\mathbf{a}^\star$ as the box that results from transforming $Q_{\mathbf{u},\mathbf{a}^\star}^{(t\rightsquigarrow{}0)}$ back into LiDAR space using the (fixed, pre-calibrated) transformation $S_C^L \in \mathbb{SE}(3)$. (Note that the pseudo-ground truth boxes must be in the same space as the predicted ones to compute the loss.)

\begin{table}[!t]
	\centering
	\vspace{2mm}
	\scriptsize
	\resizebox{\linewidth}{!}{%
		\begin{tabular}{cclccccccc}
			\textbf{Classes} & \textbf{Eval.} & \textbf{Method} & \multicolumn{7}{c}{\textbf{Accuracy@IoU}} \\
			&&& \emph{0.1} & \emph{0.2} & \emph{0.3} & \emph{0.4} & \emph{0.5} & \emph{0.6} & \emph{0.7} \\
			\midrule
			\multirow{5}{*}{Vehicle}    & 2D & TCR$^\star$ & 0.76 & 0.73 & 0.71 & 0.67 & 0.59 & 0.49 & 0.29 \\
			                                && Ours & \textbf{0.91} & \textbf{0.88} & \textbf{0.84} & \textbf{0.81} & \textbf{0.75} & \textbf{0.67} & \textbf{0.47} \\
			\cmidrule{2-10}
			                           & BEV & TCR$^\star$ & 0.69 & 0.68 & 0.64 & 0.57 & 0.50 & 0.35 & 0.12 \\
			                                && Ours & \textbf{0.86} & \textbf{0.86} & \textbf{0.84} & \textbf{0.82} & \textbf{0.77} & \textbf{0.69} & \textbf{0.39} \\
			\midrule
			\multirow{5}{*}{Pedestrian} & 2D & TCR$^\star$ & 0.28 & 0.08 & 0.00 & 0.00 & 0.00 & 0.00 & 0.00 \\
			                                && Ours & \textbf{0.48} & \textbf{0.25} & \textbf{0.09} & \textbf{0.01} & 0.00 & 0.00 & 0.00 \\
			\cmidrule{2-10}
			                           & BEV & TCR$^\star$ & 0.00 & 0.00 & 0.00 & 0.00 & 0.00 & 0.00 & 0.00 \\
			                                && Ours & \textbf{0.28} & \textbf{0.20} & \textbf{0.13} & \textbf{0.08} & \textbf{0.04} & \textbf{0.01} & \textbf{0.01} \\
			\midrule
			\multirow{5}{*}{Cyclist}    & 2D & TCR$^\star$ & 0.00 & 0.00 & 0.00 & 0.00 & 0.00 & 0.00 & 0.00 \\
			                                && Ours & \textbf{0.90} & \textbf{0.90} & \textbf{0.90} & \textbf{0.90} & \textbf{0.30} & 0.00 & 0.00 \\
			\cmidrule{2-10}
			                           & BEV & TCR$^\star$ & 0.00 & 0.00 & 0.00 & 0.00 & 0.00 & 0.00 & 0.00 \\
			                                && Ours & \textbf{0.90} & \textbf{0.90} & \textbf{0.60} & \textbf{0.10} & 0.00 & 0.00 & 0.00 \\
			\bottomrule
		\end{tabular}
	}
	\vspace{\baselineskip}
	\caption{Comparing the per-class accuracies of our method to those of TCR$^\star$ (the improved version of TCR \cite{harley2021track} published on GitHub) on the KITTI tracking benchmark \cite{geiger2012we}.}
	\label{tab:class_acc}
	\vspace{-\baselineskip}
\end{table}

\begin{figure*}[!t]
	\centering
	\includegraphics[width=\linewidth]{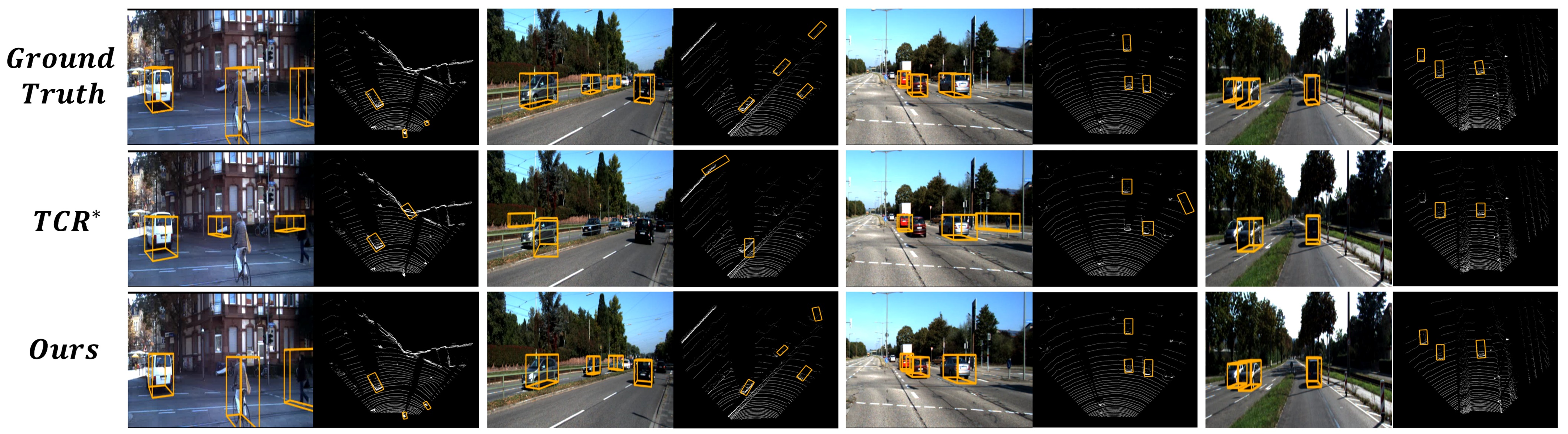}
	\caption{Qualitatively comparing our method to the state-of-the-art TCR$^\star$ method (the improved version of TCR \cite{harley2021track}) for self-supervised mobile object detection. See main text.}
	\label{fig:detection}
	\vspace{-\baselineskip}
\end{figure*}

\subsection{Loss Formulation}
\label{subsec:loss}

Our loss $L^{(t)}$ for frame $t$ can be calculated as a sum of terms, each of which compares an individual predicted box $\tilde{\mathbf{b}}_\mathbf{u}^{(t)} = (\tilde{\bm{\delta}}_\mathbf{u}^{(t)}, \tilde{\mathbf{d}}_\mathbf{u}^{(t)}, \tilde{r}_\mathbf{u}^{(t)}, \tilde{\kappa}_\mathbf{u}^{(t)})$ with its corresponding fitted box $\mathbf{b}_\mathbf{u}^{(t)} = (\mathbf{c}_\mathbf{u}^{(t)}, \mathbf{d}_\mathbf{u}^{(t)}, r_\mathbf{u}^{(t)}, \kappa_\mathbf{u}^{(t)})$. As in \S\ref{Sec:overview}, recall that the centre $\tilde{\mathbf{c}}_\mathbf{u}^{(t)}$ of $\tilde{\mathbf{b}}_\mathbf{u}^{(t)}$ can be computed from $\tilde{\bm{\delta}}_\mathbf{u}^{(t)}$, the predicted offset from the centre of the pillar corresponding to $\mathbf{u}$. We distinguish between pixels in $U_+^{(t)}$ and $U_-^{(t)}$, supervising only the confidence scores of the latter. Specifically, we compute
\begin{equation}
\scriptsize
\begin{aligned}
L^{(t)} = \sum_{\mathbf{u} \in U_+^{(t)}} & \left( \mathit{BL}_1(\tilde{\mathbf{c}}_\mathbf{u}^{(t)}, \mathbf{c}_\mathbf{u}^{(t)}) + \mathit{BL}_1(\tilde{\mathbf{d}}_\mathbf{u}^{(t)}, \mathbf{d}_\mathbf{u}^{(t)}) + \mathit{BL}_1(\tilde{r}_\mathbf{u}^{(t)}, r_\mathbf{u}^{(t)}) \; + \right. \\
& \;\; \left. \left\| \tilde{\kappa}_\mathbf{u}^{(t)} - \kappa_\mathbf{u}^{(t)} \right\|_2 \right) + \sum_{\mathbf{u} \in U_-^{(t)}} \left\| \tilde{\kappa}_\mathbf{u}^{(t)} - \kappa_\mathbf{u}^{(t)} \right\|_2,
\end{aligned}
\end{equation}
in which $\mathit{BL}_1$ denotes the balanced $L_1$ loss from \cite{pang2019libra}.

\section{Experiments}

Like TCR~\cite{harley2021track}, we experiment on the KITTI tracking benchmark~\cite{geiger2012we}, and use sequences 0000-0009 for training, and 0010 and 0011 for testing. We train our model for $300$ epochs, using Adam optimisation with an initial learning rate of $10^{-3}$ and cosine learning rate annealing. We then compare with three recent self-supervised methods that can detect objects in 2D or 3D without requiring ground-truth annotations:
(i) Slot Attention~\cite{locatello2020object} uses reconstruction error to learn a multi-block representation of the scene that can be used to detect objects within it;
(ii) SCALOR~\cite{Jiang*2020SCALOR:} is a generative tracking model that learns to capture object representations from video and enables the detection of a large number of objects in a complex scene;
(iii) TCR~\cite{harley2021track} first tries to find candidate regions using RANSAC, then, throughout multiple training stages, uses 2D-3D correspondence scores for possible object regions as a supervision signal.
To clarify the current performance drop of state-of-the-art self-supervised methods in comparison to annotated methods, we also compare to a recent semi-supervised method \cite{wang20213dioumatch} and a recent fully supervised one \cite{shi2021pv}.

\subsection{Object Discovery}

Our first set of experiments compares our method's ability to discover mobile objects to that of the other methods we consider. We evaluate both the 3D bounding boxes predicted by each method (which we compare to the 3D KITTI ground truth boxes) and their projections into the 2D camera images (which we compare to the 2D KITTI ground truth). We refer to the first of these evaluation types as BEV, and the second as 2D, in line with \cite{harley2021track}. To project each predicted 3D bounding box into 2D, we project its $8$ vertices individually into the image plane, and then use the minimum and maximum components of the projected vertices to fit the tightest possible 2D axis-aligned bounding box around them.

As shown in Table~\ref{tab:obj_discovery}, our method significantly outperforms the state-of-the-art self-supervised method TCR$^\star$ (the improved version of TCR \cite{harley2021track} published on GitHub) in both the 2D and BEV settings. It is also interesting to compare our performance to the semi-supervised 3DIoUMatch \cite{wang20213dioumatch} and the fully supervised PV-RCNN++ \cite{shi2021pv}. For PV-RCNN++, the model is trained with all of the $16,648$ annotated boxes available in our training set (sequences 0000-0009 of the KITTI tracking benchmark \cite{geiger2012we}). For 3DIoUMatch, we instead randomly select $1,664$ (i.e.\ $\approx 10\%$) of these annotated boxes for training. For IoUs up to $0.5$, our BEV mAP is within $20\%$ of that of 3DIoUMatch \cite{wang20213dioumatch} and $30\%$ of PV-RCNN++ \cite{shi2021pv}, which is encouraging given that our method is unable to leverage any ground-truth annotations. Moreover, whilst both do greatly outperform us at an IoU of $0.7$, our BEV mAP in this case is still over $3 \times$ higher than that of the previous state-of-the-art TCR$^\star$ \cite{harley2021track}.

\subsection{Per-Class Accuracies}

Our second set of experiments compares our method's per-class accuracies (at different IoUs) to those of the state-of-the-art self-supervised TCR$^\star$ approach (see Table~\ref{tab:class_acc}). Both methods are self-supervised and do not predict class labels for the objects they detect; however, we can still calculate per-class accuracies by associating each predicted box with the (labelled) ground-truth box with which it has the greatest IoU. Both methods achieve much higher accuracies for vehicles than for pedestrians/cyclists, at all IoU thresholds. However, the accuracies achieved by our method for pedestrians/cyclists are also much higher than those of TCR$^\star$, which almost completely fails for these classes. We hypothesise that this may be partly a result of the non-rigid motions exhibited by pedestrians/cyclists, which can be hard to group together without additional cues. Moreover, the pedestrians/cyclists are smaller than the vehicles and appear much less frequently in the KITTI training set, making these classes harder for networks to learn. Nevertheless, as our method takes advantage of additional cues in the form of expected object sizes and box-level consistency scores, it is able to achieve at least moderate accuracies even on these more difficult classes (see also Figure~\ref{fig:detection}).

\section{Conclusion}

In this paper, we propose SCT, a method for self-supervised mobile object detection in urban driving scenes. SCT uses additional cues such as the expected sizes of target objects to estimate pseudo-ground truth boxes, and outputs a possible box for each pixel to improve object discovery. Experimentally, our object discovery results show improvements (for all IoUs) of more than $36\%$ in BEV mAP and more than $25\%$ in 2D mAP over the state-of-the-art self-supervised TCR$^\star$ method (the improved version of TCR \cite{harley2021track} published on GitHub) on the KITTI tracking benchmark \cite{geiger2012we}. Indeed, for IoUs up to $0.5$, we achieve BEV mAP scores that are within $30\%$ of the fully supervised PV-RCNN++ \cite{shi2021pv} method, without requiring any manual annotation. Furthermore, unlike TCR$^\star$, our method is able to cope at least to some extent with trickier classes such as pedestrians and cyclists, although further work is needed to improve the accuracies for these classes at high IoUs.

\newpage

\bibliographystyle{plain}
\bibliography{egbib}

\begin{thebibliography}{10}

\bibitem{arnold2019survey}
Eduardo Arnold, Omar~Y Al-Jarrah, Mehrdad Dianati, Saber Fallah, David Oxtoby,
  and Alex Mouzakitis.
\newblock {A Survey on 3D Object Detection Methods for Autonomous Driving
  Applications}.
\newblock {\em IEEE Transactions on Intelligent Transportation Systems},
  20(10):3782--3795, 2019.

\bibitem{bao2022discovering}
Zhipeng Bao*, Pavel Tokmakov*, Allan Jabri, Yu-Xiong Wang, Adrien Gaidon, and
  Martial Hebert.
\newblock {Discovering Objects that Can Move}.
\newblock In {\em {CVPR}}, 2022.

\bibitem{Bro10c}
Thomas Brox and Jitendra Malik.
\newblock {Object Segmentation by Long Term Analysis of Point Trajectories}.
\newblock In {\em {ECCV}}, 2010.

\bibitem{iso_2015_7250_3}
BS-EN-ISO.
\newblock {Basic human body measurements for technological design: Worldwide
  and regional design ranges for use in product standards}.
\newblock {\em British Standards Institute}, 7250-3, 2015.

\bibitem{burgess2019monet}
Christopher~P Burgess, Loic Matthey, Nicholas Watters, Rishabh Kabra, Irina
  Higgins, Matt Botvinick, and Alexander Lerchner.
\newblock {MONet: Unsupervised Scene Decomposition and Representation}.
\newblock {\em arXiv:1901.11390}, 2019.

\bibitem{butler2012sintel}
Daniel~J Butler, Jonas Wulff, Garrett~B Stanley, and Michael~J Black.
\newblock {A Naturalistic Open Source Movie for Optical Flow Evaluation}.
\newblock In {\em {ECCV}}, 2012.

\bibitem{chen2021moving}
Xieyuanli Chen, Shijie Li, Benedikt Mersch, Louis Wiesmann, J{\"u}rgen Gall,
  Jens Behley, and Cyrill Stachniss.
\newblock {Moving Object Segmentation in 3D LiDAR Data: A Learning-Based
  Approach Exploiting Sequential Data}.
\newblock {\em {RA-L}}, 6(4):6529--6536, 2021.

\bibitem{choe2020evaluating}
Junsuk Choe*, Seong~Joon Oh*, Seungho Lee, Sanghyuk Chun, Zeynep Akata, and
  Hyunjung Shim.
\newblock {Evaluating Weakly Supervised Object Localization Methods Right}.
\newblock In {\em {CVPR}}, 2020.

\bibitem{choe2019attention}
Junsuk Choe and Hyunjung Shim.
\newblock {Attention-based Dropout Layer for Weakly Supervised Object
  Localization}.
\newblock In {\em {CVPR}}, 2019.

\bibitem{duan2019centernet}
Kaiwen Duan, Song Bai, Lingxi Xie, Honggang Qi, Qingming Huang, and Qi~Tian.
\newblock {CenterNet: Keypoint Triplets for Object Detection}.
\newblock In {\em {ICCV}}, 2019.

\bibitem{engelcke2019genesis}
Martin Engelcke, Adam~R Kosiorek, Oiwi~Parker Jones, and Ingmar Posner.
\newblock {Genesis: Generative Scene Inference and Sampling with Object-Centric
  Latent Representations}.
\newblock In {\em {ICLR}}, 2020.

\bibitem{eslami2016attend}
SM~Ali Eslami, Nicolas Heess, Theophane Weber, Yuval Tassa, David Szepesvari,
  Koray Kavukcuoglu, and Geoffrey~E Hinton.
\newblock {Attend, Infer, Repeat: Fast Scene Understanding with Generative
  Models}.
\newblock In {\em {NeurIPS}}, 2016.

\bibitem{kiadaki2012V}
Katerina Fragkiadaki, Geng Zhang, and Jianbo Shi.
\newblock {Video Segmentation by Tracing Discontinuities in a Trajectory
  Embedding}.
\newblock In {\em {CVPR}}, 2012.

\bibitem{gao2021ts}
Wei Gao, Fang Wan, Xingjia Pan, Zhiliang Peng, Qi~Tian, Zhenjun Han, Bolei
  Zhou, and Qixiang Ye.
\newblock {TS-CAM: Token Semantic Coupled Attention Map for Weakly Supervised
  Object Localization}.
\newblock In {\em {ICCV}}, 2021.

\bibitem{geiger2012we}
Andreas Geiger, Philip Lenz, and Raquel Urtasun.
\newblock {Are we ready for Autonomous Driving? The KITTI Vision Benchmark
  Suite}.
\newblock In {\em {CVPR}}, 2012.

\bibitem{greff2019multi}
Klaus Greff, Rapha{\"e}l~Lopez Kaufman, Rishabh Kabra, Nick Watters, Chris
  Burgess, Daniel Zoran, Loic Matthey, Matthew Botvinick, and Alexander
  Lerchner.
\newblock {Multi-Object Representation Learning with Iterative Variational
  Inference}.
\newblock In {\em {ICML}}, 2019.

\bibitem{harley2021track}
Adam~W Harley, Yiming Zuo*, Jing Wen*, Ayush Mangal, Shubhankar Potdar, Ritwick
  Chaudhry, and Katerina Fragkiadaki.
\newblock {Track, Check, Repeat: An EM Approach to Unsupervised Tracking}.
\newblock In {\em {CVPR}}, 2021.

\bibitem{he2016deep}
Kaiming He, Xiangyu Zhang, Shaoqing Ren, and Jian Sun.
\newblock {Deep Residual Learning for Image Recognition}.
\newblock In {\em {CVPR}}, 2016.

\bibitem{Jiang*2020SCALOR:}
Jindong Jiang*, Sepehr Janghorbani*, Gerard de~Melo, and Sungjin Ahn.
\newblock {SCALOR: Generative World Models with Scalable Object
  Representations}.
\newblock In {\em {ICLR}}, 2020.

\bibitem{lin2017feature}
Tsung-Yi Lin, Piotr Doll{\'a}r, Ross Girshick, Kaiming He, Bharath Hariharan,
  and Serge Belongie.
\newblock {Feature Pyramid Networks for Object Detection}.
\newblock In {\em {CVPR}}, 2017.

\bibitem{locatello2020object}
Francesco Locatello*, Dirk Weissenborn, Thomas Unterthiner, Aravindh Mahendran,
  Georg Heigold, Jakob Uszkoreit, Alexey Dosovitskiy, and Thomas Kipf*.
\newblock {Object-Centric Learning with Slot Attention}.
\newblock In {\em {NeurIPS}}, 2020.

\bibitem{bestbikelock}
The Best~Bike Lock.
\newblock {What size shed do I need for my bikes?}
\newblock Available online at
  \url{https://thebestbikelock.com/bike-storage-ideas/best-bike-storage-shed/what-size-shed-for-bikes}
  (as of 12th September 2022).

\bibitem{RobotCarDatasetIJRR}
Will Maddern, Geoff Pascoe, Chris Linegar, and Paul Newman.
\newblock {1 Year, 1000km: The Oxford RobotCar Dataset}.
\newblock {\em {IJRR}}, 36(1):3--15, 2017.

\bibitem{nimblefins}
{NimbleFins}.
\newblock {What Are The Average Dimensions Of A Car In The UK?}
\newblock Available online at
  \url{https://www.nimblefins.co.uk/cheap-car-insurance/average-car-dimensions}
  (as of 12th September 2022).

\bibitem{pan2021unveiling}
Xingjia Pan*, Yingguo Gao*, Zhiwen Lin*, Fan Tang, Weiming Dong, Haolei Yuan,
  Feiyue Huang, and Changsheng Xu.
\newblock {Unveiling the Potential of Structure Preserving for Weakly
  Supervised Object Localization}.
\newblock In {\em {CVPR}}, 2021.

\bibitem{pang2019libra}
Jiangmiao Pang, Kai Chen, Jianping Shi, Huajun Feng, Wanli Ouyang, and Dahua
  Lin.
\newblock {Libra R-CNN: Towards Balanced Learning for Object Detection}.
\newblock In {\em {CVPR}}, 2019.

\bibitem{russell2006using}
Bryan~C Russell, Alexei~A Efros, Josef Sivic, William~T Freeman, and Andrew
  Zisserman.
\newblock {Using Multiple Segmentations to Discover Objects and their Extent in
  Image Collections}.
\newblock In {\em {CVPR}}, 2006.

\bibitem{shi2021pv}
Shaoshuai Shi, Li~Jiang, Jiajun Deng, Zhe Wang, Chaoxu Guo, Jianping Shi,
  Xiaogang Wang, and Hongsheng Li.
\newblock {PV-RCNN++: Point-Voxel Feature Set Abstraction With Local Vector
  Representation for 3D Object Detection}.
\newblock {\em arXiv:2102.00463}, 2021.

\bibitem{sivic2005discovering}
Josef Sivic, Bryan~C Russell, Alexei~A Efros, Andrew Zisserman, and William~T
  Freeman.
\newblock {Discovering objects and their location in images}.
\newblock In {\em {ICCV}}, 2005.

\bibitem{teed2020raft}
Zachary Teed and Jia Deng.
\newblock {RAFT: Recurrent All-Pairs Field Transforms for Optical Flow}.
\newblock In {\em {ECCV}}, 2020.

\bibitem{wang20213dioumatch}
He~Wang, Yezhen Cong, Or~Litany, Yue Gao, and Leonidas~J Guibas.
\newblock {3DIoUMatch: Leveraging IoU Prediction for Semi-Supervised 3D Object
  Detection}.
\newblock In {\em {CVPR}}, 2021.

\bibitem{weber2000towards}
Markus Weber, Max Welling, and Pietro Perona.
\newblock {Towards Automatic Discovery of Object Categories}.
\newblock In {\em {CVPR}}, 2000.

\bibitem{wu2020deep}
Yutian Wu, Yueyu Wang, Shuwei Zhang, and Harutoshi Ogai.
\newblock {Deep 3D Object Detection Networks Using LiDAR data: A Review}.
\newblock {\em IEEE Sensors Journal}, 21(2):1152--1171, 2020.

\bibitem{xue2019danet}
Haolan Xue, Chang Liu, Fang Wan, Jianbin Jiao, Xiangyang Ji, and Qixiang Ye.
\newblock {DANet: Divergent Activation for Weakly Supervised Object
  Localization}.
\newblock In {\em {ICCV}}, 2019.

\bibitem{zhang2020inter}
Xiaolin Zhang, Yunchao Wei, and Yi~Yang.
\newblock {Inter-Image Communication for Weakly Supervised Localization}.
\newblock In {\em {ECCV}}, 2020.

\bibitem{zhu2019learning}
Yi~Zhu, Yanzhao Zhou, Huijuan Xu, Qixiang Ye, David Doermann, and Jianbin Jiao.
\newblock {Learning Instance Activation Maps for Weakly Supervised Instance
  Segmentation}.
\newblock In {\em {CVPR}}, 2019.

\end{thebibliography}

\end{document}